\documentclass{vgtc} 
\newcommand{\subscriptsize}{\fontsize{8}{9}\selectfont}

\graphicspath{{figures/}{pictures/}{images/}{./}} 

\usepackage{array}                      
\usepackage{multirow}                  
\usepackage{booktabs}                  

\newcolumntype{L}[1]{>{\raggedright\arraybackslash}p{#1}} 

\usepackage{enumitem}

\setlist[itemize]{leftmargin=*, itemsep=0pt, topsep=0pt, partopsep=0pt, parsep=0pt}
\setlist[enumerate]{leftmargin=*, itemsep=0pt, topsep=0pt, partopsep=0pt, parsep=0pt}
\setlist[description]{leftmargin=*, itemsep=0pt, topsep=0pt, partopsep=0pt, parsep=0pt}

\usepackage{amsmath}

\usepackage{times}                     

\usepackage{tabu}                      
\usepackage{lipsum}                    
\usepackage{mwe}                       
\usepackage{mathptmx}                  

\vgtccategory{Research}
\vgtcinsertpkg

\title{Does visualization help AI understand data?}

\author{
  Victoria R. Li\thanks{Equal contribution, e-mails: \{vrli, jlsun\}@college.harvard.edu}\\{\scriptsize Harvard University}
  \and
  Johnathan L. Sun\footnotemark[1]\\{\scriptsize Harvard University}
  \and
  Martin Wattenberg\thanks{e-mail: wattenberg@g.harvard.edu}\\{\scriptsize Harvard University, Google Research}
}

\teaser{
  \centering
  \includegraphics[width=\linewidth]{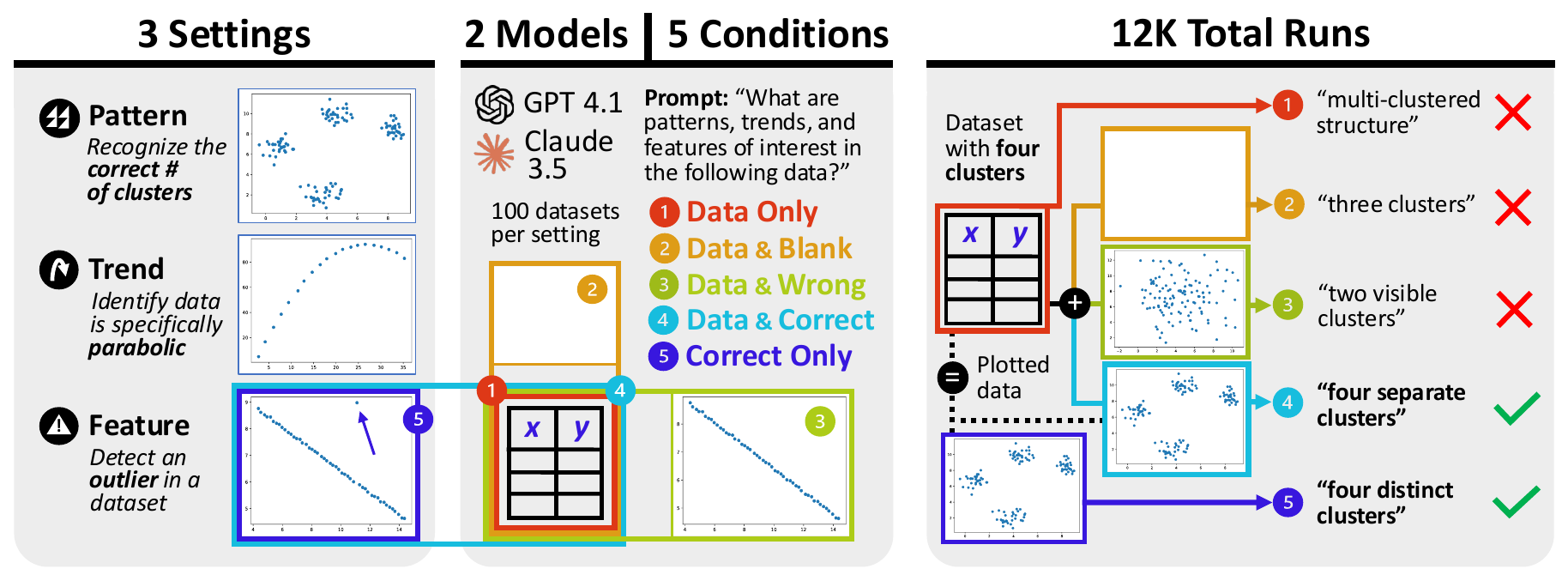}
  \caption{We test two prominent vision-language models, GPT 4.1 and Claude 3.5, on three settings motivated by common data analysis tasks. In each setting, we ask models to describe datasets under five conditions, providing: (1) numerical data alone, (2,3) numerical data with incorrect or blank visuals as baselines, (4) numerical data with correct visuals, and (5) correct visuals alone. After 12,000 trials, we find consistent performance gains when models are provided with accurate visualizations.}
  \label{fig:vis-abstract}
}

\abstract{Charts and graphs help people analyze data, but can they also be useful to AI systems? To investigate this question, we perform a series of experiments with two commercial vision-language models: GPT 4.1 and Claude 3.5. Across three representative analysis tasks, the two systems describe synthetic datasets more precisely and accurately when raw data is accompanied by a scatterplot, especially as datasets grow in complexity. Comparison with two baselines---providing a blank chart and a chart with mismatched data---shows that the improved performance is due to the content of the charts. Our results are initial evidence that AI systems, like humans, can benefit from visualization.}


\begin{document}

\maketitle

\section{Introduction}
\firstsection

For humans, visualization has long been essential for understanding our world's large and high-dimensional datasets \cite{1532781}. The recent incorporation of multi-modal artificial intelligence (AI) into question-answering and chat interfaces has created a new medium through which people explore complex data \cite{inala2024dataanalysiseragenerative}. Widespread adoption of large vision-language models (LVLMs) in these emerging workflows raises the question: \textit{does visualization also help LVLMs with dataset analysis?}  

Although LVLMs excel at many evaluation benchmarks, they often struggle with unfamiliar data distributions and tasks \cite{fang2024largelanguagemodelsllmstabular}. Nevertheless, as their capabilities and application scaffolding improve, LVLMs are increasingly used for data and visual analysis \cite{agarwal2025reviewlargelanguagemodels,inala2024dataanalysiseragenerative,xu2024exploringcapabilityllmsperforming}. AI systems are also increasingly ``agentic''---entrusted to complete tasks autonomously. To drive insight generation across diverse application areas, such as scientific discovery, models must identify dataset features that may warrant further investigation \cite{gridach2025agenticaiscientificdiscovery}.

Improving data analysis skills would enhance models' world comprehension abilities, both when deployed autonomously and as partners for human users. Moreover, if visualization proves useful for AI data analysis, researchers could open a new line of study focused on effective chart and graph design for an AI audience. 

This paper presents an initial empirical investigation into the question of whether visualization can aid AI systems. Across three common data analysis tasks, we find that visualization helps two widely adopted LVLMs---OpenAI's GPT 4.1 \cite{openai_gpt41_2025} and Anthropic's Claude 3.5 Sonnet \cite{anthropic2024claude35}---develop more accurate and specific dataset descriptions. For each task, we prompt models to identify trends, patterns, or features of interest in a synthetic setting under five conditions: (1) numerical data alone, (2--4) raw data with a blank image, misleading visualization, and correct data plot, and (5) the correct visualization alone (\cref{fig:vis-abstract}). Similarities in model behavior indicate that certain settings are well-suited to visualization-aided AI data analysis, while differences emphasize that visual designs could accommodate model idiosyncrasies. We find: 

\begin{itemize}
    \item Providing a correct visualization, with or without raw data, helps GPT and Claude accurately identify (1) the number of clusters (\cref{sec:clustering}), (2) parabolic trend (\cref{sec:nonlinear}), and (3) potential outliers (\cref{sec:outlier}) in datasets, outperforming conditions with data alone, data and a blank image, or data with an incorrect visualization.
    \item Models' performance gain given a correct plot grows with subtler tasks requiring more precise analysis, while inaccurate visualizations consistently impair model accuracy.
    \item When shown only the visualization, models generate concise responses focused on salient dataset features; given raw data, models often compute dataset summary statistics (\cref{sec:scale}).
\end{itemize}

\maketitle

\section{Related work}

\paragraph{LVLM Benchmarks} There exist many datasets for assessing LVLM chart understanding. For example, \cite{akhtar-etal-2023-reading,kantharaj-etal-2022-chart,masry2022chartqabenchmarkquestionanswering,tang-etal-2023-vistext} use tens of thousands of graphics to test models' chart question-answering  and summarization abilities. These evaluation tasks are often based on real-world examples where models are vulnerable to hallucination and training data leakage---analyzing an attached image not only based on its visual content but also unrelated domain knowledge. Different LVLMs also exhibit wide-ranging and faulty performances on these benchmarks, indicating that optimizing AI graphical understanding remains an open problem \cite{anthropic2024claude35,islam2024largevisionlanguagemodels,openai_gpt41_2025}.

In these studies, full numerical datasets, which are available in real world data analysis settings, are often excluded from model inputs. Since LVLMs can perform statistical analyses, assessing their understanding of raw datasets in realistic contexts, especially in settings without data leakage issues, is under-explored \cite{liu-etal-2024-llms}.

\paragraph{LVLMS and Visual Design} Previous work has also investigated the visual literacy of LVLMs, including the impact of graphic design elements on their understanding of underlying datasets. \cite{10670574} finds that LVLMs recognize plot titles, axes, and high-level trends, but are vulnerable to deceptive chart design or low-contrast color schemes.  \cite{mukhopadhyay2024unravelingtruthvlmsreally} shows that LVLMs perform poorly with stacked or stair plots; \cite{rahmanzadehgervi2025visionlanguagemodelsblind} describes how reducing space between points or words can lead to sharp drop-offs in LVLM visual understanding. \cite{guo2024understandinggraphicalperceptiondata} replicates a classic human visualization experiment judging chart proportions \cite{cleveland1984graphical,10.1145/1753326.1753357}, finding similarities between human and LVLM graphical perceptions for particular tasks and plot designs. These studies focus on understanding LVLM responses to visual representations without comparing them to data-only conditions.

\section{Methods}

\subsection{Three Synthetic Dataset Settings}

We base our experiments on synthetically generated data to easily vary key parameters and avoid data contamination. Our data represent three classic real-world tasks: cluster detection, identification of nonlinear trends, and outlier detection. Each task is split into four subtlety levels, and 100 datasets are independently sampled for each of the 12 task-subtlety combinations. (\cref{fig:data-examples}). Full reproduction code is provided in the supplemental materials.

\begin{figure}[!h]
    \centering
    \includegraphics[width=\linewidth]{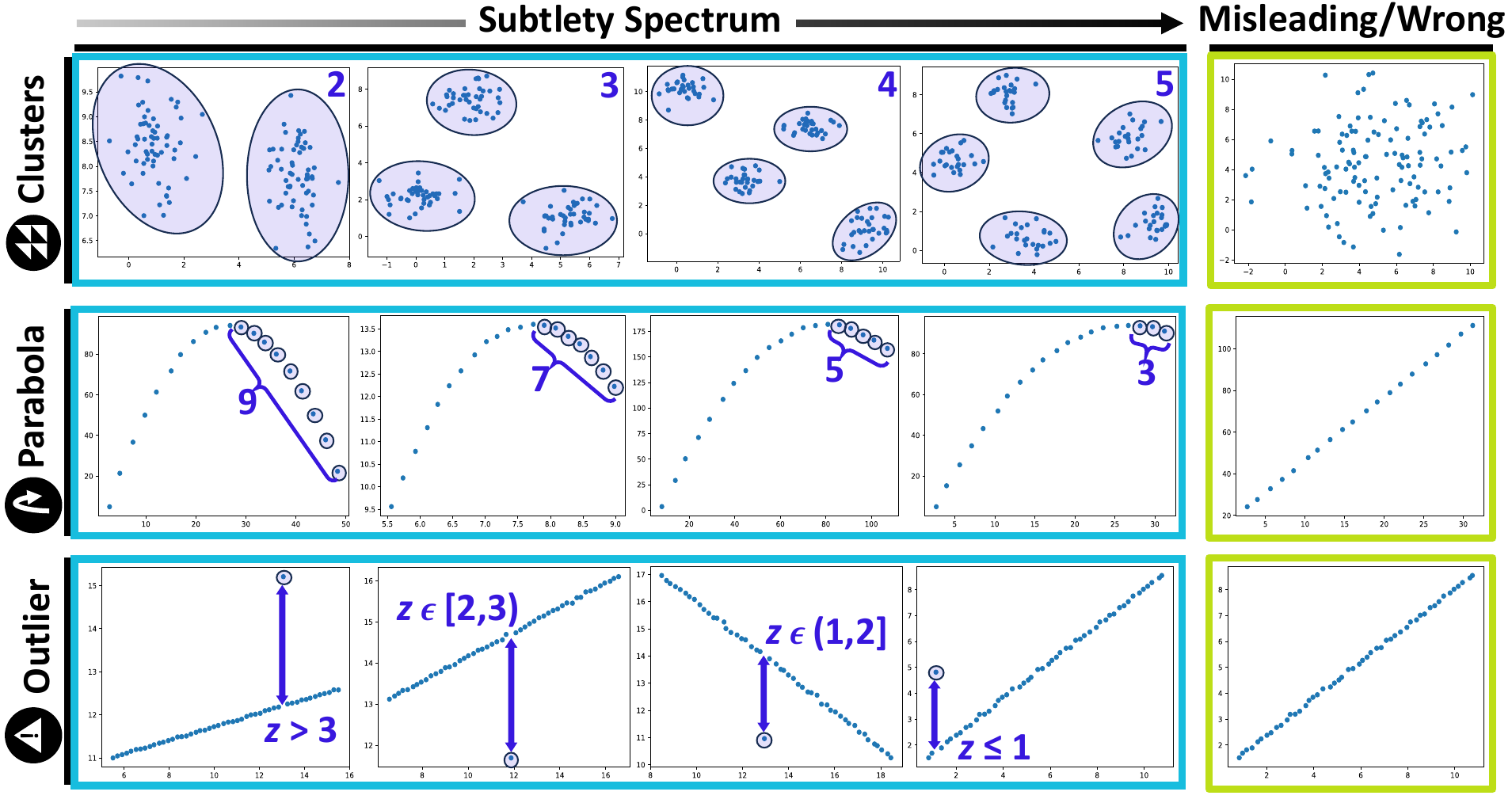}
    \caption{\textbf{The three dataset types and their subtlety spectrum.} (1) Subtler cluster datasets exhibit more groupings, requiring precise identification. Misleading ``wrong'' visuals feature one Gaussian cluster. (2) Subtler parabola datasets contain fewer points beyond the vertex, while wrong visuals show linear best-fits. (3) Subtler outlier datasets have high vertical $z$-scores (distance from mean scaled by  standard deviation), while wrong visuals remedy anomalous outliers.}
    \label{fig:data-examples}
\end{figure}

\paragraph{Clusters} Datasets include 120 points evenly distributed among 2 to 5 cluster centers. Center coordinates are sampled uniformly from the range $[0,10]$, and, points around the centers are generated as Gaussians with standard deviation $\sigma = 0.6$. To ensure distinct clusters, we discard datasets with points from different clusters less than 2 units apart. Dataset subtlety increases with more clusters. 

Misleading visuals feature a single 120-point Gaussian cluster, centered at the mean of the original cluster centers, with $\sigma$ equal to half the maximum distance from this mean to those centers.

\paragraph{Parabola} These datasets contain 20 points generated along a parabolic curve defined as $y = ax^2 + bx$. Coefficients $a$ and $b$ are uniformly drawn from $[0, 1]$ and $[1, 10]$, respectively. We then randomly select a number $n \in {3, 5, 7, 9}$ to determine how many points appear to the right of the parabola’s vertex. Points are evenly spaced along the x-axis. Gaussian noise and a random offset between 0 and 10 are added to both coordinates. Dataset subtlety increases as $n$ decreases, making the parabolic trend less clear.

Misleading visuals are the linear best-fit of the generated data, sampled at identical $x$ values, with the same added Gaussian noise.

\paragraph{Outlier} These datasets contain 50 points. Two endpoints $(0, y_1)$ and $(10, y_2)$ are randomly selected, with $y_1, y_2$ drawn from $[0, 10]$. A line connecting these endpoints is then sampled at 20 equally spaced points along the x-axis before Gaussian noise proportional to the line’s slope is added to the y-coordinates, $\mathbf{y}$. To create an outlier, one randomly selected point is vertically shifted to produce a jackknife residual greater than 2.5. Dataset subtlety increases as the outlier's vertical $z$ score, defined as $(y - \bar{\mathbf{y}})/s_{\mathbf{y}}$ (where $\bar{y}$ and $s_{\mathbf{y}}$ are the mean and standard deviation of $\mathbf{y}$, respectively), decreases.

Misleading visuals depict the data prior to outlier perturbation.

\subsection{Five Model Input Conditions}

We prompt two models to identify patterns, trends, and features of interest in data via their APIs: GPT 4.1 (\texttt{gpt-4.1-2025-04-14}) and Claude 3.5 (\texttt{claude-3-5-sonnet-20241022}) (\cref{fig:vis-abstract}). Generated synthetic datasets are randomly shuffled and provided to the models as space-separated values rounded to four significant figures. All data visualizations are default Matplotlib scatterplots, 6.4 by 4.8 inches with blue points on automatically rescaled, unlabeled axes (\cref{fig:data-examples}) \cite{Hunter:2007}. We encode plots at 300 dots-per-inch resolution in base64 for API input. For each model, setting, and subtlety level, we evaluate five experimental conditions: 

\begin{enumerate}[leftmargin=1.5em]
    \item \textbf{Data Only:} Just the raw dataset as a baseline condition.
    \item  \textbf{Data \& Blank:} An all-white image controlling for visual input itself separate from visual information impacting performance. 
    \item \textbf{Data \& Wrong:} Misleading visualization to test whether discrepant visual information degrades LVLM judgments.
    \item \textbf{Data \& Correct:} Including the plot displaying the datapoints.
    \item \textbf{Correct Only:} Scatterplot alone to assess visual understanding. 
\end{enumerate}

\begin{table}[!h]
\centering\scriptsize
\setlength{\tabcolsep}{3pt}
\renewcommand{\arraystretch}{0.9}
\begin{tabu} to \linewidth {X[2.11,l]|X[8,l]}
\toprule
\textbf{Setting} & \textbf{Response Classification} \\ \midrule
\textbf{Clustering} {\subscriptsize[LVLM Judge]} &
A Gemini judge determines how many clusters, if any, have been explicitly determined or enumerated.
\\ \midrule
\textbf{Parabola} {\subscriptsize[Keyword]} &
A string matcher looks only for `quadratic,' `parabola,' `parabolic,' `concave,' or `opens down.' It excludes broader terms like `polynomial,' `curved' or `nonlinear.' 
\\ \midrule
\textbf{Outlier} \hspace{4pt}{\subscriptsize[LVLM Judge]} &
A Gemini judge verifies that a provided correct anomalous point is explicitly flagged as an outlier. 
\\ \bottomrule
\end{tabu} \vspace{0.8em}
\caption{\textbf{Criteria for judging if an LVLM response is a ``success.''} We classify if responses (1) enumerate clusters, (2) detect a parabolic trend, and (3) identify the correct outlier.}
\label{tab:scoring-criteria}
\end{table} \vspace{-0.5em}

\paragraph{Response Classification} We use keyword search-based evaluation for one task---parabola detection---because responses largely use specific terms to describe a trend as parabolic. In contrast, the other two tasks require more nuanced evaluation: determining the exact number of clusters mentioned and if a particular point was identified as an outlier. As a result, we use a third separately developed LVLM (\texttt{gemini-2.5-flash-06-17}) for output classification in the clustering and outlier tasks. ${\sim}10$\% of evaluations were manually verified (\cref{tab:scoring-criteria}).

\section{Results}

Relative to baselines with full raw data, including a correct visualization improves GPT and Claude's dataset understanding. Across settings, performance uplift grows with dataset subtlety. 

\subsection{Clustering  Pattern}\label{sec:clustering}

\begin{figure}[!h]
    \centering
    \includegraphics[width=\linewidth]{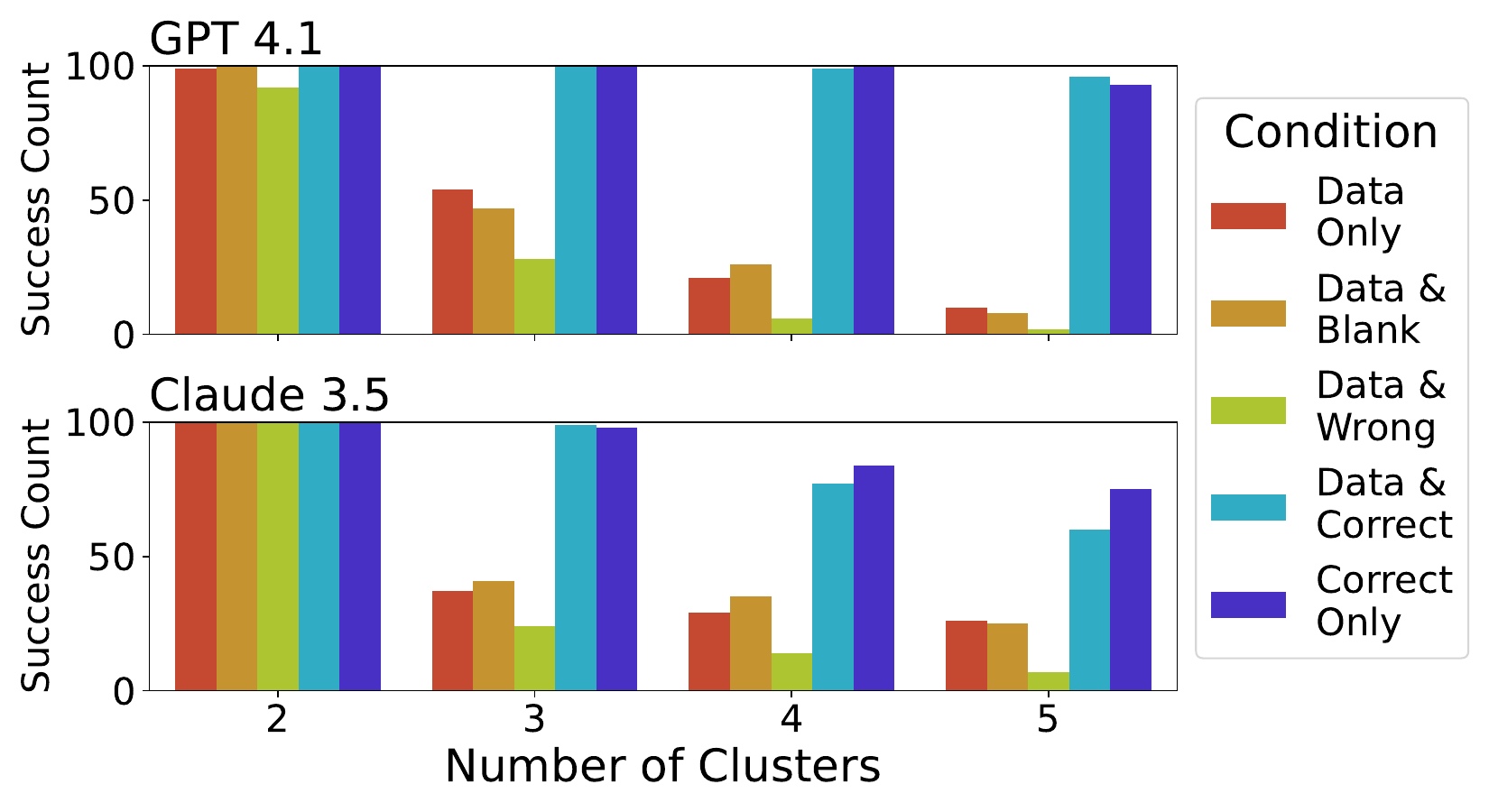}
    \caption{\textbf{Visualization improves GPT 4.1 and Claude 3.5's ability to detect the number of clusters in input datasets}. Success counts are shown for datasets containing two to five clusters. GPT consistently shows significant benefits from accurate visuals, while Claude benefits less when datasets are more subtle.}
    \label{fig:clustering_plot}
\end{figure}

Providing accurate data visualizations improves GPT and Claude's ability to correctly identify the number of clusters in datasets. Since introducing blank or misleading visuals does not improve performance relative to the baseline ``Data Only'' condition,  performance gains in conditions with correct scatterplots are attributable to models recognizing clusters via the visual (\cref{fig:clustering_plot}).

The benefit of visualization is more dramatic when datasets become more subtle and cluster identification requires more precision. With 2-cluster data, GPT improves slightly with accurate visualizations, while Claude identifies 2 clusters across conditions. However, for 4- or 5-cluster datasets, accurate scatterplots substantially boost performance: GPT and Claude are at least 2.3 and 1.7 times more accurate, respectively, with than without the correct plot.

This performance gap is driven by a sharp decline in cluster identification accuracy in the ``Data Only,'' ``Data \& Blank,'' and ``Data \& Wrong'' conditions between the 2- and 3-cluster dataset settings. This drop-off indicates LVLM dataset comprehension based solely on raw numerical inputs can be brittle. Claude's performance with correct visuals also declines as the number of dataset clusters increases from 3 to 5. Performance in the ``Data Only'' condition largely plateaus in this interval, indicating that changes in numerical and visual dataset reasoning abilities may not be strongly correlated for certain tasks and models.

Across all conditions, both models almost always fail by underestimating the number of clusters in datasets. Moreover, when it fails, Claude misidentifies the number of clusters in 5-cluster datasets by more than one in 51\% of responses with the data alone compared to 8\% of responses with correct visuals, suggesting that even when absolute accuracy declines, visuals may benefit data analysis by reducing the magnitude of LVLM errors.

Conversely, including misleading visualizations consistently leads to the worst performance across conditions for both models. Although the incorrect visual displays one cluster, in this case, both models tend to output 2 regardless of subtlety level. Evaluating responses with an LVLM, we also classify if responses mention that the included visualization disagrees with the provided data. Combined, GPT and Claude only mention a discrepancy in 7 out of 800 deceptive trials, indicating that models may silently fail, impacted by both numeric and visual inputs, given mismatches in its prompt.

\subsection{Parabolic Trend}\label{sec:nonlinear}

\begin{figure}[!h]
    \centering
    \includegraphics[width=\linewidth]{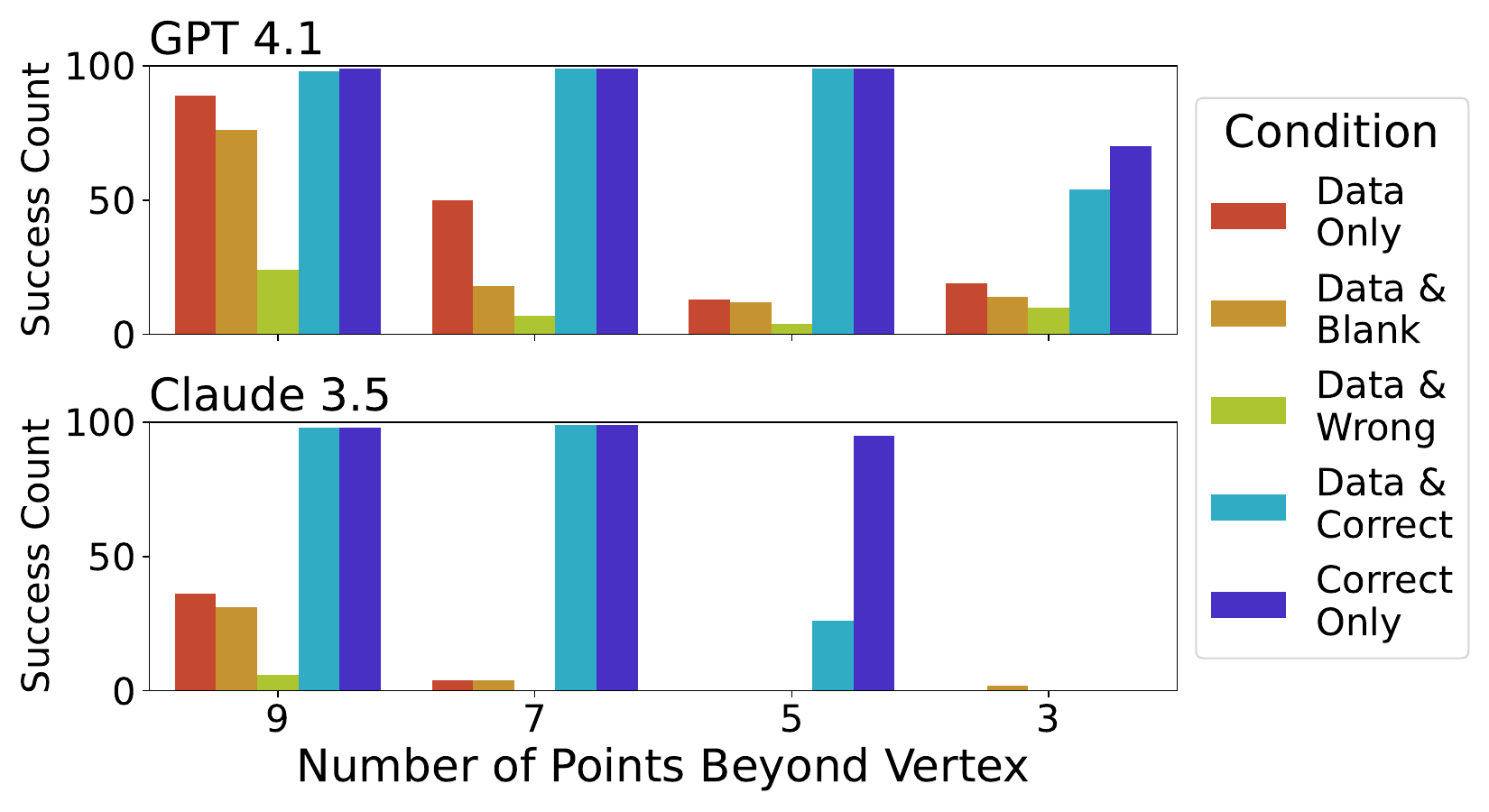}
    \caption{\textbf{Visualization improves GPT 4.1 and Claude 3.5's ability to identify parabolic trends in input datasets}. Success counts are presented for scenarios with 3 to 9 points sampled beyond the vertex. At higher subtlety levels, providing only an accurate plot results in better performance for both models than providing the data and plot.} 
    \label{fig:nonlinearity_plot}
\end{figure}

Similar to the cluster counting task, Claude and GPT perform best in parabola identification with an accurate plot, especially as dataset subtlety increases. Again, the ``Data \& Wrong'' condition consistently induces the worst performance (\cref{fig:clustering_plot,fig:nonlinearity_plot}). This trend is particularly pronounced for parabola datasets: across subtlety levels, including a misleading best-fit linear scatterplot reduces success counts by more than half compared to any other condition.

Since this is a substantial effect, we further analyze the impact of misleading linear visualizations on parabolic datasets. Across models, broad trend descriptors like ``nonlinear'' or ``curved'' are common in the ``Data Only'' and ``Data \& Blank'' conditions, but their frequency decreases up to eightfold with a misleading visual. This finding underscores the significant influence visualization could have on AI---including a graphic can override a model's interpretation and discussion of raw data. 

There are also notable differences between the clustering and parabolic settings. For all cluster datasets, models achieve non-zero success counts, but for parabolic trend identification, Claude has no meaningful performance for the most subtle 3-points-beyond-the-vertex dataset setting. At this level, Claude mentions ``logarithmic,'' ``logistic,'' or ``sigmoid'' in 25\% of responses with the data alone but in 91\% of responses with the correct visual, revealing that models can also misinterpret correct visuals.

Moreover, while models showed minimal performance differences between the ``Data \& Correct'' and ``Correct Only'' conditions for clustering datasets (\cref{fig:clustering_plot}), for the most subtle parabolic datasets, the correct visualization-only condition outperforms all others. For instance, when Claude is presented with datasets containing 5 points beyond a parabola's vertex, including data alongside the correct scatterplot reduces performance by a factor of 3.7. GPT demonstrates a small effect in the same direction on datasets containing 3 points beyond the vertex. Determining the scenarios where visualizations alone provide more useful input relative to raw data could guide better design of AI-assisted data analysis tools.

\subsection{Outlier Feature}\label{sec:outlier}

\begin{figure}[!h]
    \centering
    \includegraphics[width=\linewidth]{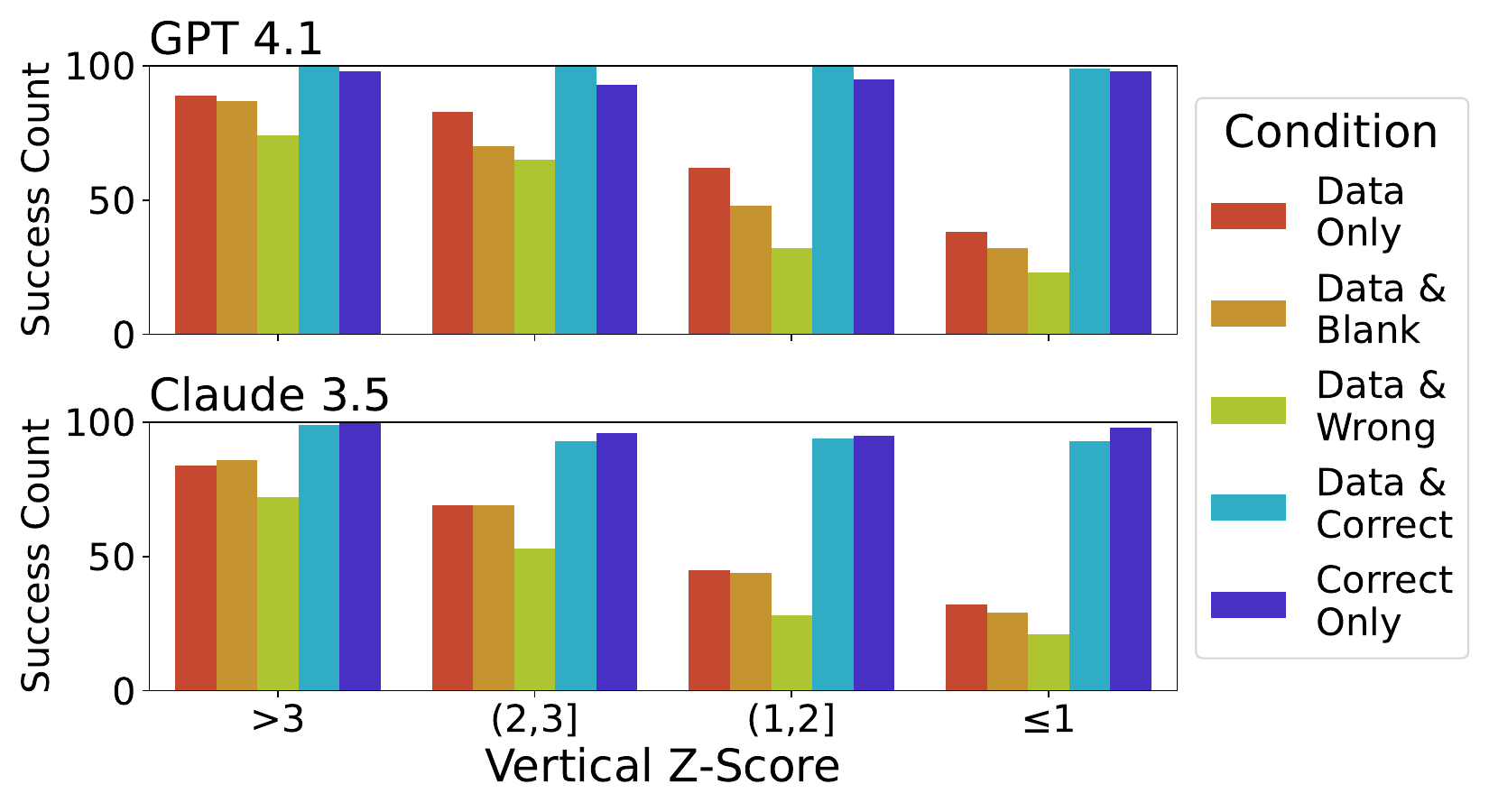}
    \caption{\textbf{Visualization improves GPT 4.1 and Claude 3.5's ability to identify the correct outlier in input datasets}. Success counts are presented for scenarios with binned ${\leq}1$ to ${>}3$ vertical $z$-scores. As subtlety increases, both models' performance gradually decline, especially without accurate visuals.} 
    \label{fig:oulier-plt}
\end{figure}

Finally, visualization helps both models identify a specific outlier in linear datasets, with the benefit of visualization increasing with task subtlety. The misleading ``Data \& Wrong'' condition also remains worse than other conditions (\cref{fig:oulier-plt}). 

Unlike the previous two settings, both GPT and Claude generally perform comparably or better when provided both the dataset and the correct plot compared to only the correct plot. This trend may have to do with the characteristics of the task: outlier detection requires the retrieval and analysis of specific data values rather than a feature involving multiple datapoints (like clustering) or the entire dataset (like the parabolic trend).

\subsection{General Trends}
\label{sec:scale}

\begin{figure}[!h]
    \centering
    \includegraphics[width=\linewidth]{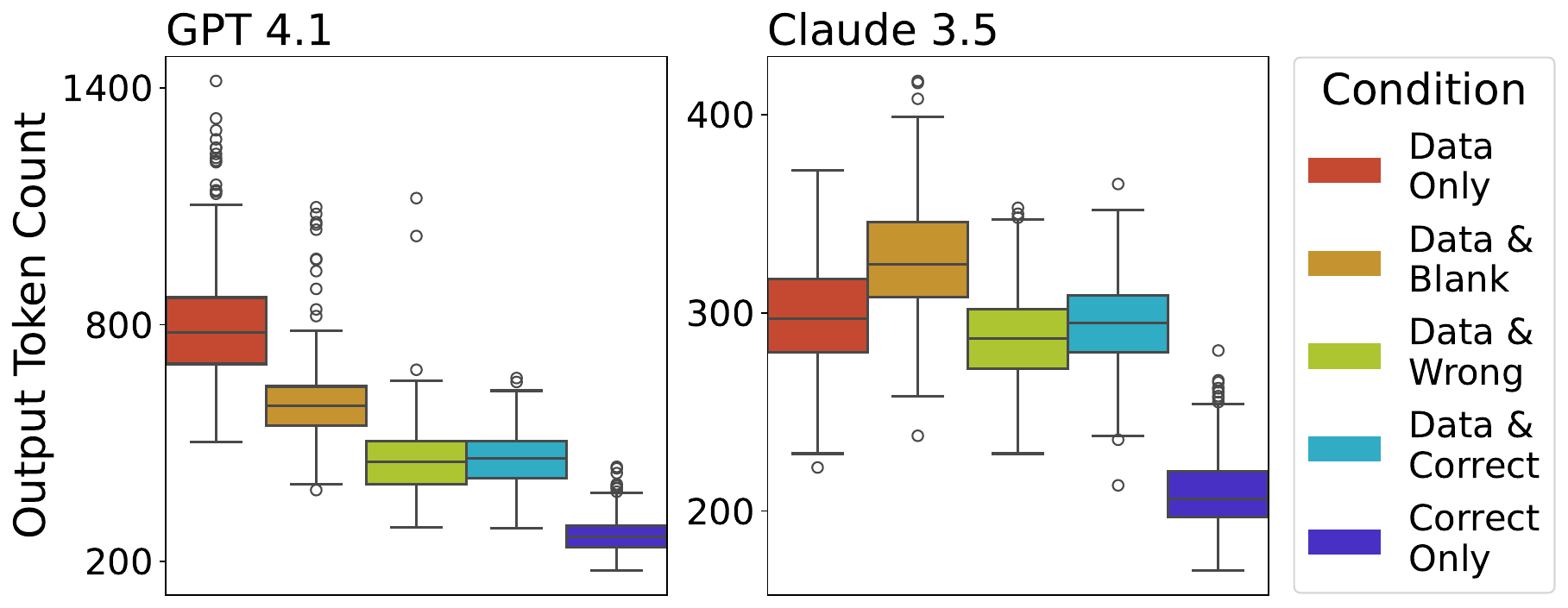}
    \caption{\textbf{Representative output token counts in the outlier detection setting}. Without a visual, GPT often outputs the entire dataset before analyzing it. For both GPT and Claude, providing only the correct visual leads to the most concise answers.}
    \label{fig:token-counts}
\end{figure}

We also summarize some overall observations across settings. First, model responses vary greatly in length. On average across all datasets and conditions, Claude responses are 288 tokens long while GPT responses are 884 tokens long. One reason for this discrepancy is that without an accurate visual, GPT often outputs the entire data table before any analysis. That said, GPT may use these tokens productively to reason about the data. Also, across all settings, more recently-released GPT-4.1's performance generally matches and sometimes significantly exceeds Claude-3.5's performance (\cref{fig:clustering_plot,fig:nonlinearity_plot,fig:oulier-plt}). This result agrees with existing findings on vision-language benchmarks like MMMU and Mathvista, and indicates visualization can help models across capability levels \cite{openai_gpt41_2025,anthropic2024claude35}.

Furthermore, both models generate the shortest responses
when provided with a visual alone. Any visual input, including a blank image, decreases the length of GPT but not Claude's output length. In the high performance scatterplot-only condition, both models generate brief responses focused on salient dataset features (\cref{fig:token-counts}). 

In contrast, models often compute summary statistics in their responses when provided with only data. We calculate the frequency of terms in responses referring to the domain, range, mean, min, max, and variance of the data. Across settings, these summary
statistic terms appear two times more often on average in GPT responses with the data alone compared to visuals alone, and up to ten times more often on average in Claude responses.

\section{Discussion and Conclusion}

Our findings provide evidence that visualization can help AI systems understand data. This fact raises tantalizing questions that lie beyond the scope of this short paper. How do graphical variations on our charts affect the results? After all, a large part of the field of visualization focuses on the efficacy of different charts and visual encodings---perhaps there is an entire parallel version of that body of scholarship for AI systems. If it turns out the same graphical ideas are best for both humans and AI, that raises the intriguing possibility that interpreting AI systems could shed light on human psychology; on the other hand, if there are systematic differences, we may need to build up a new field of AI-oriented visualization.

Our results are primarily a proof-of-concept. We use simple tasks, only evaluate two models, and leverage fully synthetic datasets. This set up demands follow-up work with additional tasks, visualization types, and LVLMs to substantiate our findings. Nevertheless, the large effects we observe suggest our results may be generalizable across workflows and models, opening a new area of visualization research. We conclude by summarizing our findings organized around questions we hope will inspire follow-up work:

\textbf{``Who''} --- which AI models benefit most from visualization in analytical tasks?  We did not use AI with cutting-edge tool use or inference-time compute capabilities. Future work could confirm whether a larger sample of models with varied application scaffolding still benefits from visualization. Crucially, visualizations designed for AI should also consider the requirements of human users to ensure that AI-generated insights remain interpretable, transparent, and useful for practical decision-making. 

\textbf{What} are the best metrics to evaluate dataset understanding? While we do evaluate whether models complete our analytical tasks, there is significant room for additional validation, such as confirming the logical consistency of AI-generated insights. Employing LVLM judges similar to ours is one scalable method to characterize model responses. Future studies should also continue investigating how models handle misleading or inconsistent visuals, which we show can have a large effect on dataset analysis.

\textbf{When} is visualization most beneficial? Across all three settings, visualization benefits models most when datasets are subtle and challenging to precisely interpret. Further research may test the boundaries of these findings, generalizing to new datasets or more nuanced analytical tasks.

\textbf{Where} does AI currently struggle with visualization? We observed instances where Claude misinterprets parabolic plots as sinusoidal or logistic. Addressing this limitation could entail gaining deeper insights into how models internally reconcile visual and numerical information, which could also help better integrate them into data-processing pipelines. 

\textbf{Why} is visualization helpful? Visualization is essential for humans partly because our visual processing systems efficiently synthesize complex information. Advanced AI models, despite learning about our world through fundamentally different mechanisms, similarly benefit from visual aids. Exploring these parallels could deepen our understanding of the underlying principles by which visualizations effectively convey information, facilitate pattern recognition, and enable knowledge discovery. 

\textbf{How} can future research optimize visualizations for AIs? Given that AI models are abundant and testable, future work could explore methodologies tailored to AI model interpretation by systematically searching over optimal visualization parameters. These models can be prompted thousands of times at relatively low cost, potentially allowing for personalized visualization design.

Overall, our work lays groundwork for exciting avenues of research at the intersection of visualization and AI, suggesting that visualization could be a broad tool for emerging human, AI, and human-AI workflows.

\section{Acknowledgements}

The authors thank Fernanda Vi\'egas,  Catherine Yeh, Shivam Raval, and Andrew Zhang for helpful advice and insightful discussions. The authors also gratefully acknowledge API credit support from AISST, particularly Ryan Kaufman. MW is supported by the Effective Ventures Foundation, Effektiv Spenden Schweiz, a Superalignment grant from OpenAI, and the Open Philanthropy Project. 

\subsection{Supplementary Materials}

This work's supplemental materials, including code and additional figures, are available at \href{https://github.com/johnathansun/lvlm-vis-data-understanding}{github.com/johnathansun/lvlm-vis-data-understanding}. 

\bibliographystyle{abbrv-doi}
\bibliography{main}
\end{document}